\pdfoutput=1

\documentclass[11pt]{article}
\usepackage{authblk}

\usepackage[final]{acl}

\usepackage{times}
\usepackage{latexsym}

\usepackage[T1]{fontenc}

\usepackage[utf8]{inputenc}

\usepackage{microtype}

\usepackage{inconsolata}

\usepackage{graphicx}
\usepackage{amsmath}
\usepackage{amsthm}
\usepackage{booktabs}
\usepackage{algorithm}
\usepackage{algorithmic}
\usepackage[switch]{lineno}
\usepackage{multirow}
\usepackage{pdfpages}
\usepackage{tabularx}
\usepackage{xcolor}
\usepackage{xurl}

\newcommand{\redtext}[1]{\textcolor{black}{#1}}

\title{Context-Aware Sentiment Forecasting via LLM-based Multi-Perspective Role-Playing Agents}

\author{
    \textbf{Fanhang Man}\textsuperscript{1},
    \textbf{Huandong Wang}\textsuperscript{2},
    \textbf{Jianjie Fang}\textsuperscript{3},
    \textbf{Zhaoyi Deng}\textsuperscript{4}, \\
    \textbf{Baining Zhao}\textsuperscript{1},
    \textbf{Xinlei Chen}\textsuperscript{1}\thanks{Corresponding Author},
    \textbf{Yong Li}\textsuperscript{2}
    \\
    \textsuperscript{1}Shenzhen International Graduate School, Tsinghua University,\\
    \textsuperscript{2}Department of Electronic Engineering, Tsinghua University,\\
    \textsuperscript{3}Northeastern University at Qinghuangdao,\\
    \textsuperscript{4}Department of Computer Science, University of California, Irvine
    \\
    
    \texttt{\href{mailto:mfh21@mails.tsinghua.edu.cn}{mfh21@mails.tsinghua.edu.cn},\quad\href{mailto:chen.xinlei@sz.tsinghua.edu.cn}{chen.xinlei@sz.tsinghua.edu.cn}}
}

\begin{document}
\maketitle
\begin{abstract}
User sentiment on social media reveals the underlying social trends, crises, and needs. Researchers have analyzed users' past messages to trace the evolution of sentiments and reconstruct sentiment dynamics. However, predicting the imminent sentiment of an ongoing event is rarely studied. In this paper, we address the problem of \textbf{sentiment forecasting} on social media to predict the user's future sentiment in response to the development of the event. We extract sentiment-related features to enhance the modeling skill and propose a multi-perspective role-playing framework to simulate the process of human response. Our preliminary results show significant improvement in sentiment forecasting on both microscopic and macroscopic levels.
\end{abstract}
\section{Introduction}
The study of sentiments on social media has facilitated social research~\cite{levy2022understanding}, marketing~\cite{zhang2021mining}, and public management~\cite{solovev2022moral}. For instance, during public emergencies, measuring negative sentiments can assist disaster relief organizations in adjusting their key objectives to focus on areas with heightened collective distress~\cite{zhang2020does}. The task of \textbf{sentiment forecasting} timely anticipates the sentiment of a person or a crowd with the available information on social media. While distinct from sentiment analysis and other related tasks, forecasting complements these approaches by offering forward-looking insights that enhance our understanding of sentiment dynamics and enable proactive decision-making..

Retrospective sentiment analysis methods typically assign a discrete or linear sentiment score to sentences, paragraphs, or entire documents~\cite{mousavi2022effective}. Researchers leveraged deep neural networks~\cite{hu2018multimodal}, Transformer~\cite{zhong2021useradapter}, and Bert~\cite{islam2022ar} to extract users' sentiment states. The evolution of sentiments is also heavily studied~\cite{tu2022viral}. \citet{Okawa2022PredictingOD} leveraged a sociologically informed neural network (SINN) to deduce the users' sentiments as they evolve. ~\citet{Liu2022APL} integrated the DeGroot Model with a probabilistic linguistic method to simulate people's sentiments. However, these methods only consider the reciprocal influences among social media users. In the real-world scenario, the nature of human sentiment is highly context-dependent~\cite{kuppens2017emotion, halim2020machine, hu2018multimodal}. The development of ongoing social events highly affects social media users' sentiments. However, the ongoing events have complex semantic information, which is too intricate to be formulated as an input for the above methods. 


The advent of large language models (LLMs) sheds light on comprehensive and prospective sentiment reasoning in real-world scenarios~\cite{chang2023survey}. LLMs understand complex external event contexts with embedded common-sense knowledge~\cite{zha2025enable, zhao2025embodied}. \citet{zhang2023instruct} fine-tuned GPT to analyze the sentiment with rich contexts. Moreover, LLMs can capture the nuanced tone of voice in texts including sarcasm, humor, and rhetorical questions~\cite{wang2023chatgpt, safdari2023personality}. \citet{deng2023llms} applied such an advantage to analyze sentiment for Reddit. These works formalized sentiment analysis as a reasoning problem to integrate semantic information and subjectivity in human opinion~\cite{hou2024progressive}. Such advantages provide an opportunity to pivot from traditional retrospective studies of sentiment analysis to prospective studies of sentiment forecasting.


However, there are still gaps in accurately forecasting the sentiment of different users on social media. To begin with, it requires comprehensive user features to model the user behavior~\cite{iosifidis2017large}. These labels are difficult to obtain on social media due to anonymity and privacy regulations. Moreover, even with sufficient labels, the evolution of human sentiments is too intricate and subtle to model~\cite{kuppens2017emotion,wang2022systematic}. People in different circumstances would have diverse sentiment responses to the same events~\cite{hu2024resemo}. Yet it is challenging to model these subtle sentimental clues.

In this paper, we focus on the problem of \textbf{Sentiment Forecasting} to predict people's future sentiments on social media under real-world events. To comprehensively incorporate the context, we leverage LLMs to understand the complex semantic information. To enrich the features for user-specific sentiment modeling, we extract the features from user-generated content, specifically the tone of voice and attitude toward the event. To address the complexity and variety of sentiment evolution, we develop a multiperspective role-playing method to forecast the users' social media comments. Specifically, a general LLM would role-play the diverse users to express themselves on social media. A fine-tuned "psychologist" LLM with expertise would analyze the response and feedback to the general LLM for reflections to ensure consistency. The proposed method addresses the complexity of user-specific sentiment forecasting, enabling more precise and nuanced predictions of both individual and collective sentiments.

Our contributions lie in the following aspects:
\begin{itemize}
    \item We focus on \textbf{Sentiment Forecasting} on social media, aiming to predict the future sentiment response of social media users towards ongoing real-world events. We formulate it as a reasoning problem to incorporate external context.
    \item To enrich user-specific features, we implemented an LLM-based feature extraction method targeting implicit features to simulate the sentiment response of social media users.
    \item We present a multiperspective role-playing method to predict future sentiment responses. A general LLM role-plays common social media users with extracted features to generate responses. A fine-tuned LLM role-plays objective psychologist to ensure behavioral consistency.
\end{itemize}

\section{Related Work}

\subsection{The Study of Sentiment}
The study of sentiments is crucial in understanding the public opinion in diverse evets settings including public events~\cite{solovev2022moral, li2022tract}, natural disasters~\cite{li2024physics}, and livelihood~\cite{li2024quest, liu2024mobiair}.
Researchers use agent-based modeling and network theory to model the changes in collective sentiment~\cite{castellano2009statistical}. For example, \citet{Okawa2022PredictingOD} leveraged a sociologically-informed neural network to track and predict the evolution of user sentiments over time. \citet{Liu2022APL} integrated the DeGroot Model with a probabilistic linguistic method to forecast people's decisions during the COVID-19 pandemic. \cite{monti2020learning} fit the agent model with real-world social traces to recover the real-world dynamics. The above methods merely construct the evolution based on mutual interactions of the social media users without any event context, compromising applicability in real-world scenarios. \redtext{Our proposed framework adopted the development of ongoing real-world events as context and user-specific circumstances as sentiment clues to tackle the problem of timely real-world deployment.}

\subsection{Role-Play with LLMs}
LLMs exhibit sophisticated dialogue behavior~\cite{abbasiantaeb2024let}. \citet{shanahan2023role} introduced the idea of role-play to characterize the phenomena for LLMs to perform the part of a person or a superposition of simulacra within a multiverse of possible characters. Role-play methods let the LLM act as the user themselves to convey the dialogue. Several researchers therefore fine-tuned the LLMs to role-play characters from entertainment works, including animations~\cite{Li2023ChatHaruhiRA}, TV series~\cite{Zhou2023CharacterGLMCC, Wang2023RoleLLMBE}, and Movies~\cite{Chen2022LargeLM}. Others also applied such a technique to recreate notable historical figures~\cite{Zhou2023CharacterGLMCC, Shao2023CharacterLLMAT}. However, existing methods incorporated massive character-centered data, e.g., personality, social status, and relationships, to train the LLMs to role-play the character. In the online social media scenario, it is unrealistic and unethical to obtain such data from social media users. Though inspiring, such methods could only be used to predict some specific fabricated character's reaction to events with limited generalizability in real-life \redtext{social media}.

\begin{figure*}[htbp]
  \centering
  \includegraphics[width=\textwidth]{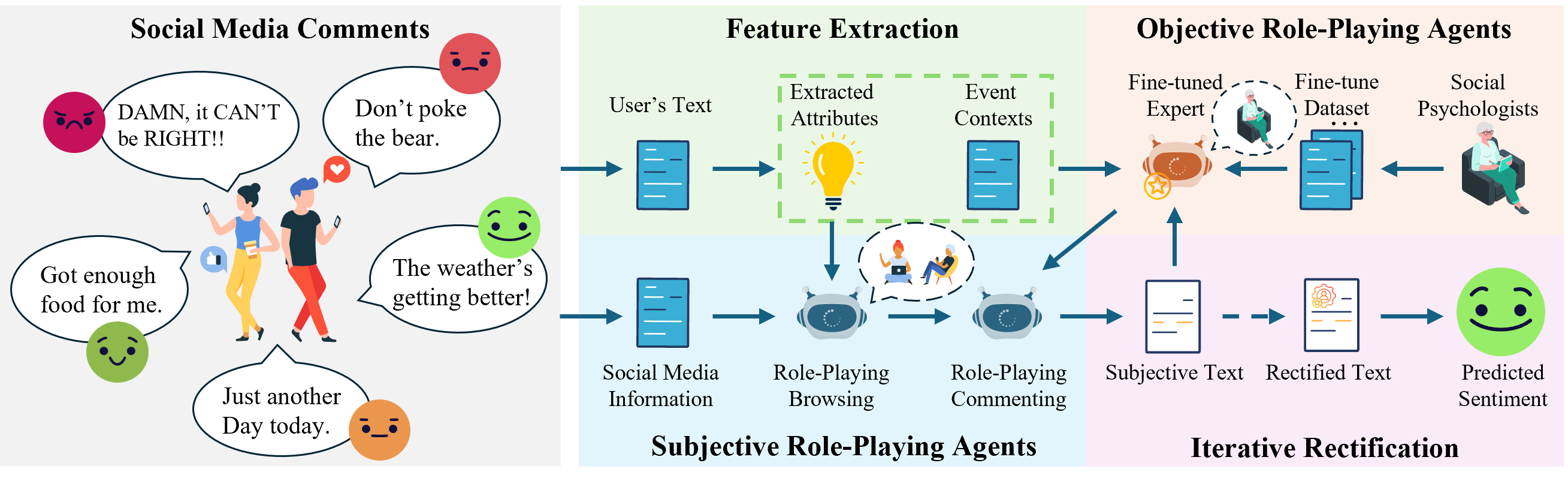} 
  \caption{Architecture of the proposed context-aware sentiment forecasting framework.} 
  \label{fig: framework}
\end{figure*}

\section{Methodology}

In this section, we first formally define the problem of sentiment forecasting. Then we outline the proposed context-aware sentiment forecasting framework (Figure \ref{fig: framework}), followed by detailed descriptions of each core component: feature extraction, subjective role-playing agent, objective role-playing agent, and iterative rectification. Formal notation definitions are summarized in Table \ref{tab: notations}. 

\subsection{Problem Formulation}
For social media, traditional sentiment analysis operates retrospectively, mapping an existing user comment $c$ to a sentiment space via:
\begin{equation} 
\label{eqn: sa}
\begin{aligned} 
\sigma=F_{\text{SA}}(c),
\end{aligned}
\end{equation}
where $F_{SA}$ denotes the sentiment analysis function.

However, sentiment forecasting constitutes a prospective temporal reasoning task that requires systematic logical inferences, including identification of historical patterns, contextual cues, and future sentiment trends. Formally, sentiment forecasting aggregates historical semantic information up until time $t$ to predict the future sentiment at the time of interest $t'$, where $t'\ge t$. With the ability to gather social media comments, sentiment forecasting can be timely conducted as the ongoing event evolves. The task of sentiment forecasting is formalized as follows: 
\begin{equation} 
\label{eqn: definition}
\begin{aligned} 
\sigma_{t'}=F_{\text{SF}}(\bigcup_{ \tau \le t}\cdot_{\tau}),
\end{aligned}
\end{equation}
where $\cdot_\tau$ indicates the information at time $\tau$ and $F_{SF}$ is the function of sentiment forecasting.

\subsection{Overall Framework}\label{subsec: framework}
Sentiment forecasting aims to conjecture the future sentiment of a social media user or crowd. To effectively integrate the semantic information from user attributes, user comments, and event context for sentiment forecasting, we develop an LLM-based multi-perspective role-playing framework. The subjective agent generates future comments where the sentiment lies. The objective agent applies knowledge in behavioral psychology to discriminate abnormality in the generated comments to restrict stochasticity.


As shown in Figure~\ref{fig: framework}, the framework comprises four components: feature extraction, subjective role-playing agent, objective role-playing agent, and iterative rectification. Feature extraction aims to identify implicit features derived from existing social media comments. This process is designed to capture the user's habitual textual tone of voice and infer the user's attitude towards the ongoing event. With the extracted features, the subjective role-play agent simulates the behavior of a social media user to comprehend the specific contexts and skim through the followees' social media comments. Subsequently, the agent is instructed to generate a new comment at the future time of interest $t'$ regarding the event. To ensure consistency in the user's textual tone of voice and attitude flow, a fine-tuned objective role-play agent serves as a behavioral psychologist to review this generated comment to filter potential behavioral inconsistency. During iterative rectification, the analysis from the objective agent is fed back to the subjective role as a guide to iteratively regenerate a rectified comment. Finally, the forecasted sentiment is retrieved with the state-of-the-art sentiment analysis method. Further details are provided in the subsections below.

\subsection{Feature Extraction}\label{subsec: feature extraction}
The feature extraction model aims to retrieve social media users' sentiment-centered elements. In addition to the self-reported user attributes $\mathcal{A}$, which includes gender, religion, location, etc., the implicit sentiment-centered aspects underlined in the social media comments are also critical. We define the set of all (attainable) social media comments of a user $u$ to be $\mathcal{C}^u = \{c^u_\tau | \tau \in T^u\}$, where $c^u_\tau$ is a specific comment made by $u$ at time $\tau$ and $T^u$ is the set of all time points at which user $u$ has made comments. Since the comment is predicted at $t' \ge t$, we are specifically interested in comments $\mathcal{C}^u_{t}= \{c^u_\tau | \tau \in T^u, \tau \le t\}$ before $t$ during the feature extraction to avoid spoiled information.

In case of social media comments, a user's intentional choice of lexical, syntactic, and paralinguistic elements reflects the unique way of expression and persona. Such a choice is defined as the \textbf{textual tone of voice} $\nu$. Studies demonstrate that people tend to maintain a consistent textual tone of voice according to their social image on social media~\cite{cingi2023tone, rettberg2017self}. The textual tone of voice is extracted via an LLM, which is formalized as follows:
\begin{equation} 
\label{eqn: tau}
\begin{aligned} 
\nu^u_{t}=\text{LLM}(\mathcal{C}^u_{t}, i_{\nu}),
\end{aligned}
\end{equation}
where $\nu^u_{t}$ indicates the extracted tone of voice with information before $t$. $i_{\nu}$ denotes the instruction of analyzing the textual tone of voice. The output is structured as three descriptive adjectives. Examples are demonstrated in Appendix \ref{app: instruction demonstration}.

In addition, attitude is a psychological construct representing an individual's enduring evaluative stance towards certain aspects. In our case, we address the user attitude toward public events. The event context $\mathcal{E}^u_{t}$, experienced or witnessed by user $u$ before $t$, serves as an important factor when inferring user attitude. Unlike the high consistency in the textual tone of voice, user attitude may reasonably shift as the event evolves. Yet the textual tone of voice highly affects the expression of user attitude. Therefore, user attitude is extracted as follows:
\begin{equation} 
\label{eqn: alpha}
\begin{aligned} 
\alpha^u_t=\text{LLM}(\nu^u_t, \mathcal{C}^u_{t},\mathcal{E}^u_{t}, i_{\alpha}),
\end{aligned}
\end{equation}
where $\alpha^u_t$ represents user attitude of user $u$ before time $t$. $i_{\alpha}$ represents the designed instruction to analyze user attitude toward the event. 

\begin{table}
    \centering
    \begin{tabularx}{\columnwidth}{Xr}
        \toprule
        Symbol   & Description of Notations                      \\
        \midrule
        $u$                   & Social media user indicator      \\         
        $\mathcal{C}^u$       & Social media comments of $u$     \\
        $\mathcal{T}^u$       & The set of time points $u$ made comments \\
        $\mathcal{A}^u$       & Self-reported attributes of $u$          \\
        $\mathcal{F}^u$       & $u$'s Followees' social media comments   \\
        $\mathcal{E}^u$       & Context of the event experienced by $u$  \\
        $i$                   & Instruction for a specific task  \\
        $\nu$                 & Textual tone of voice            \\
        $\alpha$              & Attitude towards the event       \\
        $\phi$                & Predicted social media comment   \\
        $\theta$              & Behavioral psychological analysis\\                
        $\sigma$              & Sentiment indicator              \\
        \bottomrule 
    \end{tabularx}
    \caption{Definition of Notations}
    \label{tab: notations}
\end{table}

\subsection{Subjective Role-playing Agent}\label{subsec: SRA}
The subjective role-play strategy instructs the LLM to role-play a social media user based on the extracted user features. We strictly limit the input to before time $t$ to role-play the user. The role-playing LLM is instructed to maintain the textual tone of voice and avoid unreasonable attitude shifts. The user's social media comments $\mathcal{C}^u_{t}$ before $t$ are provided as few-shot learning samples. The subjective role-playing agent $\text{LLM}^s_{t}$ is formalized as:
\begin{equation}
\label{eqn: rp_s}
\begin{aligned}
\text{LLM}^s_{t} = \text{RP}(\nu^u_{t}, \alpha^u_{t}, \mathcal{A}^u_{t}, \mathcal{C}^u_{t}, \mathcal{E}^u_{t}, i_{r}),
\end{aligned}
\end{equation}
where RP indicates the role-playing process for LLM$^s_{t}$ to role-play a subjective social media user with the information before time $t$. $i_{r}$ is the instruction for role-play. $\mathcal{A}^u_{t}$ is the attainable self-reported user attribute of user $u$ on social media. 

With the role-playing LLM, we could simulate the reactive process with two stages: browsing social media and commenting. A sample of the followees' social media comments $\mathcal{F}^u_t$ is extracted based on empirical media influence factors, such as relevance to the topic, frequency of interaction, number of followers, etc. The role-playing LLM browses through these comments and predicts a comment to be sent at the time of interest $t'$. These processes can be integratively formalized as follows: 
\begin{equation} 
\label{eqn: Ts}
\begin{aligned} 
\phi^u_{t'}=\text{LLM}^s_t(\mathcal{F}^u_{t}, \mathcal{E}^u_{t}, t', i_{s}),
\end{aligned}
\end{equation}
where $\phi^u_{t'}$ is the generated future comment to be posed at time $t'$ from a user regarding the ongoing event. $i_s$ is the instruction for the act of browsing and commenting.

\subsection{Objective Role-playing Agent}\label{subsec: ORA}
To mitigate subjectiveness and prevent unusual stochastic behaviors, we fine-tune an objective psychologist LLM to critically analyze the generated comments.

\textbf{Constructing fine-tune dataset:} We first gathered ten sets of social media comments from Twitter/X. Noted that these samples are carefully gathered to avoid overlap among events during the testing. The event context spans natural disasters, political events, social events, etc. With the method detailed in Sections \ref{subsec: feature extraction} and \ref{subsec: SRA}, we extracted $\nu^u_t$ and $\alpha^u_t$ from the comments and generated the predicted social media commment $\phi^u_{t'}$. Then, we enlisted three experts in behavioral psychology. The experts were asked to independently assess whether the LLM$^s_{t}$-generated comments maintained a consistent textual tone of voice and reasonably coherent attitude flow with the original user comments $C^u_t$. Their "yes" or "no" judgments were followed by brief written analyses of observed consistencies/inconsistencies. Agreement analysis revealed a 70\% inter-expert percentage agreement and a Fleiss’ Kappa~\cite{fleiss1981measurement} of 0.796, indicating substantial agreement. Both metrics demonstrate strong consensus, supporting the reliability of our annotations.With their analyses as few-shot samples, we construct over 25,000 sets of such reviews with GPT 4o.

\textbf{Finetuning psychologist LLM:} The reviews are utilized for low-rank adaptation (LoRA) supervised fine-tuning, wherein the pre-trained model weights are kept intact and the weight matrices are modified using low-rank decomposition~\cite{hu2021lora}. This approach increases the task-specific parameter gains, embedding expert behavioral psychology knowledge into the fine-tuned model, functioning as an objective behavioral psychologist. The fine-tuning process is demonstrated as follows:
\begin{equation}
\label{eqn: ft}
\begin{aligned}
\lambda^{(m+1)} = \lambda^{(m)} - \eta \nabla_{\lambda} L(\lambda^{(m)}),
\end{aligned}
\end{equation}
where $\lambda^{(m)}$ represents the parameters at the $m$-th iteration, and $L(\lambda)$ is the loss function that measures the prediction error of the model on the given task. By computing the gradient $\nabla_{\lambda}L(\lambda^{(t)})$, we determine how the parameters should be updated to reduce the error. The learning rate $\eta$ determines the size of each update step. 
Further details of the fine-tuning process are illustrated in Appendix \ref{app: fine tuning}. 

\textbf{Consistency analysis:} The fine-tuned model is defined as $\text{LLM}^o_t$, role-playing a behavioral psychologist with event context untill time $t$. LLM$^o_t$ is assigned to analyze the textual tone of voice consistency between the LLM$^s_{t}$ generated social media comment $\phi^u_{t'}$ and the user's previous social media comment $\mathcal{C}^u_{t}$ and oversee potential unreasonable attitude shift.
\begin{equation} 
\label{eqn: analyze}
\begin{aligned} 
\theta^u_{t'}=\text{LLM}^o_t(\nu^u_{t}, \alpha^u_{t}, \mathcal{C}^u_{t}, \phi^u_{t'}, i_{o}),
\end{aligned}
\end{equation}
where $\theta^u$ is the analysis from the fine-tuned psychologist LLM, and $i_o$ is the instruction for consistency analysis.

\subsection{Iterative Rectification}
The objective LLM would determine if the generated comment demonstrates consistency. A comment that passes the consistency test is treated as the comment to be sent at time $t'$. For inconsistent comments, the analysis $\theta^u_{t‘}$ from psychologist LLM$^o_t$ is fed to the subjective role \redtext{LLM$^s$} along with its generated comment $\phi^u_{t'}$ for a more consistent comment regeneration, which is formalized as follows: 
\begin{equation}
\label{eqn: rp_r}
\begin{aligned}
\phi_{t'}^u = \text{LLM}^s_t(\theta^u_{t'}, \mathcal{A}^u_{t}, \mathcal{C}^u_{t}, \phi^u_{t'}, i_g),
\end{aligned}
\end{equation}
\redtext{where $\phi_{t'}^u$ on the left side of the equation is the regenerated social media comment for the role-played social media user whose previously generated comment is determined to be inconsistent by LLM$^o_t$. $i_g$ indicates the instruction for rectification.}

The regenerated comments would be iteratively analyzed by the psychologist LLM$^o_t$ with a limit of $n$ iterations

\begin{table*}[!t]
\centering
\caption{Macroscopic performance comparison of two datasets, evaluated using JSD. Bold indicates the best (lowest) JSD score, and underlining denotes the second-best result.}
\label{tab: macroscopic}
\resizebox{\textwidth}{!}{%
\begin{tabular}{@{}lcccccccccccccc@{}}
\toprule
Dataset& \multicolumn{8}{|c|}{2012 Hurricane Sandy} & \multicolumn{4}{c}{\redtext{2020 U.S. Election}} \\
Metrics& \multicolumn{4}{|c}{Distribution of Sentiment} & \multicolumn{4}{c|}{Distribution of Polarity} & \multicolumn{2}{c}{Dis. of Sen.} & \multicolumn{2}{c}{Dis. of Pol.}\\
Location & \multicolumn{2}{|c}{New Jersey} & \multicolumn{2}{c}{New York} & \multicolumn{2}{c}{New Jersey} & \multicolumn{2}{c|}{New York} & \multicolumn{4}{c}{/} \\
Time & \multicolumn{1}{|c}{T1} & T2 & T1 & T2 & T1 & T2 & T1 & \multicolumn{1}{c|}{T2} & T3 & T4 & T3 & T4\\
\midrule
Voter       & \multicolumn{1}{|c}{0.2822} & 0.3234 & \multicolumn{1}{|c}{0.2245} & 0.2581 & \multicolumn{1}{|c}{0.2543} & 0.2423 & \multicolumn{1}{|c}{0.2454} & 0.2341 & \multicolumn{1}{|c}{0.1289} & 0.1129 & \multicolumn{1}{|c}{0.1030} & 0.1006 \\
DeGroot     & \multicolumn{1}{|c}{0.2376} & 0.1869 & \multicolumn{1}{|c}{0.2511} & 0.2243 & \multicolumn{1}{|c}{0.2082} & 0.1744 & \multicolumn{1}{|c}{0.1310} & 0.1213 & \multicolumn{1}{|c}{0.1854} & 0.1738 & \multicolumn{1}{|c}{0.1204} & 0.1183\\
SLANT+      & \multicolumn{1}{|c}{0.3630} & 0.30355 & \multicolumn{1}{|c}{0.2912} & 0.2874 & \multicolumn{1}{|c}{0.2516} & 0.2681 & \multicolumn{1}{|c}{0.2905} & 0.2734 & \multicolumn{1}{|c}{0.1490} & 0.1408 & \multicolumn{1}{|c}{0.1331} & 0.1243\\
NN          & \multicolumn{1}{|c}{0.1904} & 0.2161 & \multicolumn{1}{|c}{0.1733} & 0.1804 & \multicolumn{1}{|c}{0.1517} & 0.1379 & \multicolumn{1}{|c}{0.1270} & 0.1355 & \multicolumn{1}{|c}{0.0482} & 0.0441 & \multicolumn{1}{|c}{0.0267} & 0.0228\\
SINN        & \multicolumn{1}{|c}{0.1673} & 0.1359 & \multicolumn{1}{|c}{0.1504} & 0.1377 & \multicolumn{1}{|c}{0.1201} & 0.1312 & \multicolumn{1}{|c}{0.1022} & 0.1216 & \multicolumn{1}{|c}{0.0554} & 0.0625 & \multicolumn{1}{|c}{0.0294} & 0.0363\\
MPR$_{G}$ & \multicolumn{1}{|c}{\underline{0.0243}} & \textbf{0.0105} & \multicolumn{1}{|c}{\underline{0.0456}} & \underline{0.0396} & \multicolumn{1}{|c}{\underline{0.0211}} & \textbf{0.0064} & \multicolumn{1}{|c}{\underline{0.0290}} & \underline{0.0239} & \multicolumn{1}{|c}{\textbf{0.0097}} & \textbf{0.0053} & \multicolumn{1}{|c}{\textbf{0.0024}} & \textbf{0.0013}\\
MPR$_{M}$ & \multicolumn{1}{|c}{\textbf{0.0148}} & \underline{0.0192} & \multicolumn{1}{|c}{\textbf{0.0220}} & \textbf{0.0313} & \multicolumn{1}{|c}{\textbf{0.0114}} & \underline{0.0100} & \multicolumn{1}{|c}{\textbf{0.0116}} & \textbf{0.0204} & \multicolumn{1}{|c}{\underline{0.0106}} & \underline{0.0068} & \multicolumn{1}{|c}{\underline{0.0038}} & \underline{0.0017}\\
\bottomrule
\end{tabular}%
}
\end{table*}

\section{Experiment}

\subsection{Experiment Settings}

\textbf{Datasets: } We conducted extensive experiments on two datasets corresponding to two large-scale socially-aware events, i.e., the 2012 Hurricane Sandy and the 2020 U.S. Presidential Election. \redtext{To help understand the course of events, brief summaries of both datasets and events are shown in Appendix~\ref{app: summary}}. Both datasets were collected from Twitter (renamed to X in 2023).

Both datasets allow in-depth analysis not only of the text itself but also of the temporal-spatial attributes, providing valuable insights into social media behaviors during the event. The sentiment label is obtained using the state-of-the-art supervised learning model developed by NLPtown, bert-base-multilingual-uncased-sentiment\footnote{\url{https://huggingface.co/nlptown/bert-base-multilingual-uncased-sentiment}}, which achieves an accuracy as high as 87\% on multiple datasets~\cite{sahoo2023comparative, contreras2023deep}. 

In addition, for each event, we picked two distinct time points to perform sentiment forecasting (T1 and T2 for 2012 Hurricane Sandy, T3 and T4 for the 2020 U.S. Election).  For the 2012 Hurricane Sandy Dataset, T1 is the time immediately after Sandy hit New Jersey on Oct. 29, 2012. T2 is one week later on Nov. 5, 2012, with significant post-disaster relief. For the 2020 U.S. Election dataset, T3 represents the time after the second presidential debate on Oct. 29, 2020. T4 represents the time after President-elect Biden claimed victory on Nov. 7, 2020. These markers are chosen to align with major developments, ensuring rich contextual relevance, and are separated by at least a week to avoid overlap and isolate distinct phases of the events.
Details of data preprocessing are elaborated in Appendix~\ref{app: preprocessing}.

\begin{figure*}
    \centering
    \includegraphics[width=\linewidth]{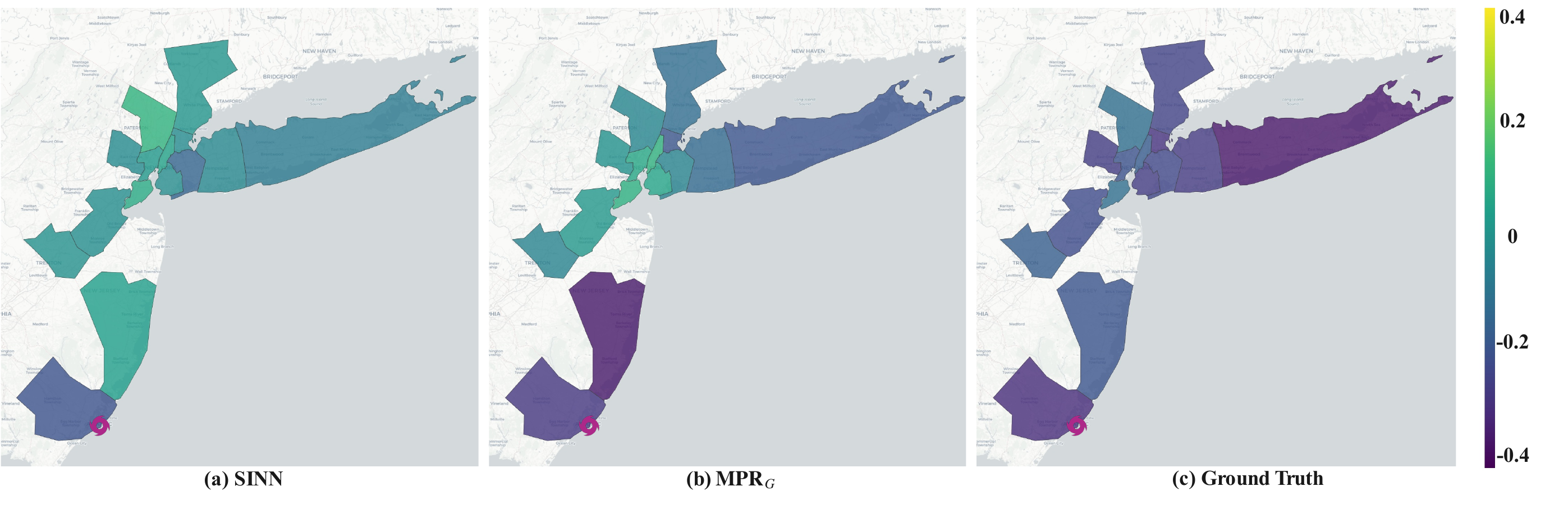}
    \caption{The collective sentiment level for selected New York and New Jersey counties. The red circle indicates the location of the Landfall of Hurricane Sandy. Figures (a) and (b) are the reconstructed sentiment distribution maps for SINN and our proposed MPR$_{G}$, respectively. Figure (c) is the ground truth distribution.}
    \label{fig: sentiment_map}
\end{figure*}

\textbf{Implementation Details: } For the subjective role-playing agent, we applied Gemma 2 9B~\cite{team2024gemma} and Mistral NeMo 12B~\cite{jiang2023mistral} to role-play the social media users and generate future comments at the time of interest. Such choices are because the popular models like the GPT series are designed not to process or generate inappropriate or offensive content, which is prevalent on social media. Moreover, testing both models shows the robustness of our proposed frameworks. For the objective role-playing agent, we leveraged Llama 3 8B Instruct. The learning rate $\eta$ is set to be $1\times10^{-4}$ to prevent acute update while maintaining an accessible rate of convergence~\cite{hu2021lora}. For iterative rectification, the limit of iterations $n$ is set to be 3 to balance the computational efficiency and the effectiveness of the module.

\noindent\textbf{Baselines: }The dynamics of information and sentiment propagation of social media have been a heavily researched topic. We compare it against state-of-the-art methods, including social model-based and neural network-based methods. (1) \textbf{Voter}  adapts the users' sentiment from their followees~\cite{muslim2024mass}. (2) \textbf{DeGroot} assumes users update their sentiment iteratively to the weighted average of the followee's sentiment~\cite{Degroot1974ReachingAC, Wu2023MixedOD}. (3) \textbf{SLANT+} is a non-linear generative model applying a recurrent neural network (RNN) with the point process model to learn the non-linear evolution of the user's sentiment~\cite{Kulkarni2017SLANTAN}. (4) \textbf{NN} is a pure neural network method to learn the evolution of users' sentiments based on previous patterns~\cite{de2016learning}. (5) \textbf{SINN} is the Sociologically-informed Neural Network model, which harnesses sociological models to guide neural networks to approach users' sentiment evolution~\cite{Okawa2022PredictingOD}.

\noindent\textbf{Metrics: }We evaluate the performance of sentiment forecasting with real-world data on both microscopic and macroscopic levels. The microscopic evaluation measures if the prediction is accurate for each individual user with accuracy and Macro F1 scores. The macroscopic evaluation focused on the distribution of a general crowd sharing similar characteristics. We adopted the Jensen-Shannon divergence (JSD) to measure the similarity between the distribution of the forecasted sentiment and the ground truths for a crowd, which can be represented as follows:
\begin{equation} 
\label{eqn: jsd}
\begin{aligned} 
\text{JSD}(\textbf{p}||\textbf{q}) = \frac{1}{2}\text{KL}(\textbf{p}||\frac{\textbf{p}+\textbf{q}}{2}) + \frac{1}{2}\text{KL}(\textbf{q}||\frac{\textbf{p}+\textbf{q}}{2}),
\end{aligned}
\end{equation}
where $\textbf{p}$ and $\textbf{q}$ are two distributions and $\text{KL}(\cdot||\cdot)$ is the Kullback-Leibler divergence. The lower the JSD, the closer the distribution between the prediction and the ground truth.

Inspired by sentiment analysis works with fine granularity, we used the distribution of sentiment and the distribution of sentiment polarity $p=\text{sgn}(s)$ to measure the fidelity of the forecast. Sentiment has five categories \{-2, -1, 0, 1, 2\} ranging from strongly negative to strongly positive. Sentiment polarity is on a scale of \{-1, 0, 1\} where only the polarity of the sentiment is considered.

\begin{table*}[!t]
\centering
\caption{The microscopic performance for different models for datasets from two different states during the landfall of Hurricane Sandy. Bold denotes the best (highest) score and underline denotes the second-best score.}
\label{tab: microscopic}
\resizebox{\textwidth}{!}{%
\begin{tabular}{@{}lcccccccccccccc@{}}
\toprule
Dataset & \multicolumn{8}{|c}{2012 Hurricane Sandy} & \multicolumn{4}{|c}{\redtext{2020 U.S. Election}} \\
Location & \multicolumn{4}{|c}{New Jersey} & \multicolumn{4}{c}{New York} & \multicolumn{4}{|c}{/}\\
Time & \multicolumn{2}{|c}{T1} & \multicolumn{2}{c}{T2} & \multicolumn{2}{c}{T1} & \multicolumn{2}{c}{T2}  & \multicolumn{2}{|c}{T3} & \multicolumn{2}{c}{T4} \\
Metrics & \multicolumn{1}{|c}{Acc.} & Ma. F1 & Accu. & Ma. F1 & Acc. & Ma. F1 & Acc. & Ma. F1 & \multicolumn{1}{|c}{Acc.} & Ma. F1 & Acc. & Ma. F1  \\
\midrule
Voter       & \multicolumn{1}{|c}{0.199} & 0.133 & \multicolumn{1}{|c}{0.169} & 0.128 & \multicolumn{1}{|c}{0.194} & 0.137 & \multicolumn{1}{|c}{0.156} & 0.125 & \multicolumn{1}{|c}{0.347} & 0.189 & \multicolumn{1}{|c}{0.387} & 0.198 \\
DeGroot     & \multicolumn{1}{|c}{0.183} & 0.143 & \multicolumn{1}{|c}{0.310} & 0.217 & \multicolumn{1}{|c}{0.187} & 0.150 & \multicolumn{1}{|c}{0.288} & 0.196 & \multicolumn{1}{|c}{0.238} & 0.185 & \multicolumn{1}{|c}{0.217} & 0.184 \\
SLANT+      & \multicolumn{1}{|c}{0.213} & 0.168 & \multicolumn{1}{|c}{0.227} & 0.187 & \multicolumn{1}{|c}{0.190} & 0.131 & \multicolumn{1}{|c}{0.202} & 0.137 & \multicolumn{1}{|c}{0.332} & 0.169 & \multicolumn{1}{|c}{0.345} & 0.170 \\
NN          & \multicolumn{1}{|c}{0.285} & 0.146 & \multicolumn{1}{|c}{0.302} & 0.155 & \multicolumn{1}{|c}{0.253} & 0.187 & \multicolumn{1}{|c}{0.239} & 0.125 & \multicolumn{1}{|c}{0.426} & 0.200 & \multicolumn{1}{|c}{0.491} & 0.186 \\
SINN        & \multicolumn{1}{|c}{0.353} & 0.179 & \multicolumn{1}{|c}{0.364} & 0.167 & \multicolumn{1}{|c}{0.385} & 0.168 & \multicolumn{1}{|c}{0.327} & 0.152 & \multicolumn{1}{|c}{0.476} & 0.193 & \multicolumn{1}{|c}{0.485} & 0.183 \\
MPR$_{G}$  & \multicolumn{1}{|c}{\underline{0.413}} & \underline{0.302} & \multicolumn{1}{|c}{\textbf{0.453}} & \textbf{0.331} & \multicolumn{1}{|c}{\underline{0.396}} & \underline{0.292} & \multicolumn{1}{|c}{\underline{0.418}} & \textbf{0.329} & \multicolumn{1}{|c}{\textbf{0.615}} & \textbf{0.374} & \multicolumn{1}{|c}{\textbf{0.596}} & \textbf{0.397}  \\
MPR$_{M}$  & \multicolumn{1}{|c}{\textbf{0.445}} & \textbf{0.312} & \multicolumn{1}{|c}{\underline{0.438}} & \underline{0.309} & \multicolumn{1}{|c}{\textbf{0.482}} & \textbf{0.310} & \multicolumn{1}{|c}{\textbf{0.429}} & \underline{0.301} & \multicolumn{1}{|c}{\underline{0.593}} & \underline{0.368} & \multicolumn{1}{|c}{\underline{0.581}} & \underline{0.370} \\
\bottomrule
\end{tabular}
}
\end{table*}

\subsection{Experiment Results}

\textbf{Macroscopic Performance:} Table \ref{tab: macroscopic} presents the performance of our models. The Gemma and Mistral variants of our multi-perspective role-play (MRP) framework are denoted as MPR$_G$ and MPR$_M$, respectively. For the macroscopic JSD metric, our proposed framework is an order of magnitude better than the baselines, showing a significantly closer distribution of sentiment w.r.t. the ground truth.

The substantial performance improvement stems from a better understanding of context and better user modeling. The model-based methods tend to converge in the long run, with a practically fixed distribution. Learning-based methods also rely heavily on past sentiment scores, especially initialization. It is practically unrealistic for a user in SINN to shift drastically from -2 to 2 or vice versa. However, it is a normal behavior on social media. Among the baselines, SINN achieves the best performance since it adopts social models, i.e. Stochastic bound confidence model, to mitigate the purely data-driven neural networks. A social model could only focus on the dynamics among the social media users, but not the environmental context, which evolves as the development of the event. The proposed MPR framework, however, considers the context information through reasoning. As shown in Table 3, our integration of context information enhances the accuracy of sentiment forecasting when events shift drastically. Hurricane Sandy's landfall significantly altered the living environment for social media users. In the case of the U.S. election, when the swing states were claimed by a president-elect, it would deterministically change the outcome, dramatically affecting people's sentiment.

Moreover, for the 2012 Hurricane Sandy dataset, we analyzed users from different geographical regions to assess group-specific performance. As shown in Figure~\ref{fig: sentiment_map}, we compared collective sentiment across 15 selected counties in New York and New Jersey. While SINN predicted more positive collective sentiments, our approach better captured the sentiment shifts before and after Hurricane Sandy's landfall, leading to more accurate collective sentiment forecasting.

\textbf{Microscopic Performance:} In addition to the crowd-level analysis, we also heavily tested the performance of our framework at an individual level. As shown in Table \ref{tab: microscopic}, our proposed framework outperformed the baselines for both Gemma2 and Mistral NeMo. With Gemma2, our proposed framework achieved an average of 6.23\% improvement in accuracy and 14.7\% improvement in Macro F1 compared to the best baseline SINN on the 2012 Hurricane Sandy Dataset. For the 2020 U.S. election dataset, the accuracy improved 12.5\% and Macro F1 improved 19.3\%.   Mistral NeMo demonstrates a slightly better overall performance, with an increase of 9.13\%, 10.7\% in accuracy and 14.15\%, 17.6\% in Macro F1, for the hurricane and election datasets, respectively. With the proposed framework, we have a 45\% chance on average to correctly forecast the users' sentiments based solely on information available from news and social media. 

Sentiment forecasting for a social media user or a group of users with more specific attributes and contextual information would significantly enhance accuracy. The prompt can be further refined with a higher granularity for specific users, allowing the event context to be tailored w.r.t. the user's circumstances. This would enable the MPR framework to generate social media comments that more closely reflect the user's actual situation.

\begin{table*}[!t]
\caption{The ablation study was performed by removing components across three models. Mistral represents an experiment performed with Mistral NeMo. Bold denotes the best (highest) results.}
\label{tab: ablation}
\resizebox{\textwidth}{!}{%
\begin{tabular}{ccccccccccc}
\hline
                                              Dataset
                                              & \multicolumn{1}{l}{}                     & \multicolumn{1}{l}{} & \multicolumn{8}{c}{2012 Hurricane Sandy} \\                                        Grainularity   & \multicolumn{1}{l}{}                     & \multicolumn{1}{l}{} & \multicolumn{4}{c}{Sentiment}                                     & \multicolumn{4}{c}{Sentiment Polarity}                                      \\ 
                                                                                       Time        &                                          &                       & \multicolumn{2}{c}{T1}                  & \multicolumn{2}{c}{T2} & \multicolumn{2}{c}{T1}                  & \multicolumn{2}{c}{T2} \\ 
                                                                                       Metrics        &                                          &                       & Accuracy & \multicolumn{1}{c}{Macro F1} & Accuracy   & Macro F1   & Accuracy & \multicolumn{1}{c}{Macro F1} & Accuracy   & Macro F1   \\ \hline
\multicolumn{1}{c}{\multirow{8}{*}{\rotatebox{90}{Gemma 2}}}                                                  & \multicolumn{1}{c}{\multirow{4}{*}{NJ}} & MPR-RP        & \multicolumn{1}{|c}{0.212}   & \multicolumn{1}{c|}{0.186}   & 0.206     & 0.197     & \multicolumn{1}{|c}{0.461}   & \multicolumn{1}{c|}{0.357}   & 0.385     & 0.346     \\
\multicolumn{1}{c}{}                                                                         & \multicolumn{1}{c}{}                    & MPR-FE                 & \multicolumn{1}{|c}{0.343}   & \multicolumn{1}{c|}{0.266}   & 0.380     & 0.285     & \multicolumn{1}{|c}{0.504}   & \multicolumn{1}{c|}{0.385}   & 0.513     & 0.415     \\
\multicolumn{1}{c}{}                                                                         & \multicolumn{1}{c}{}                    & MPR-OB                   & \multicolumn{1}{|c}{0.408}   & \multicolumn{1}{c|}{0.294}   & 0.449     & 0.323     & \multicolumn{1}{|c}{0.508}   & \multicolumn{1}{c|}{0.422}   & 0.582     & 0.476     \\
\multicolumn{1}{c}{}                                                                         & \multicolumn{1}{c}{}                    & MPR                   & \multicolumn{1}{|c}{\textbf{0.413}}   & \multicolumn{1}{c|}{\textbf{0.342}}   & \textbf{0.453}     & \textbf{0.331}     & \multicolumn{1}{|c}{\textbf{0.523}}   & \multicolumn{1}{c|}{\textbf{0.436}}   & \textbf{0.585}     & \textbf{0.477}     \\ \cline{2-11} 
\multicolumn{1}{c}{}                                                                         & \multicolumn{1}{c}{\multirow{4}{*}{NY}} & MPR-RP               & \multicolumn{1}{|c}{0.181}   & \multicolumn{1}{c|}{0.166}   & 0.173     & 0.164     & \multicolumn{1}{|c}{0.408}   & \multicolumn{1}{c|}{0.340}   & 0.475     & 0.371     \\
\multicolumn{1}{c}{}                                                                         & \multicolumn{1}{c}{}                    & MPR-FE                 & \multicolumn{1}{|c}{0.280}   & \multicolumn{1}{c|}{0.234}   & 0.392     & 0.295     & \multicolumn{1}{|c}{0.495}   & \multicolumn{1}{c|}{0.371}   & 0.526     & 0.421     \\
\multicolumn{1}{c}{}                                                                         & \multicolumn{1}{c}{}                    & MPR-OB                & \multicolumn{1}{|c}{\textbf{0.393}}   & \multicolumn{1}{c|}{0.286}   & \textbf{0.418}     & 0.307     & \multicolumn{1}{|c}{0.491}   & \multicolumn{1}{c|}{0.405}   & 0.547     & 0.447     \\
\multicolumn{1}{c}{}                                                                         & \multicolumn{1}{c}{}                    & MPR                   & \multicolumn{1}{|c}{0.380}   & \multicolumn{1}{c|}{\textbf{0.292}}   & 0.412     & \textbf{0.329}     & \multicolumn{1}{|c}{\textbf{0.492}}   & \multicolumn{1}{c|}{\textbf{0.413}}   & \textbf{0.557}     & \textbf{0.453}     \\ \hline
\multicolumn{1}{c}{\multirow{9}{*}{\begin{tabular}[c]{@{}c@{}}\rotatebox{90}{Mistral}\end{tabular}}} & \multicolumn{1}{c}{\multirow{4}{*}{NJ}} & MPR-RP            & \multicolumn{1}{|c}{0.261}   & \multicolumn{1}{c|}{0.211}   & 0.225     & 0.209     & \multicolumn{1}{|c}{0.394}   & \multicolumn{1}{c|}{0.342}   & 0.381     & 0.374     \\
\multicolumn{1}{c}{}                                                                         & \multicolumn{1}{c}{}                    & MPR-FE              & \multicolumn{1}{|c}{0.420}   & \multicolumn{1}{c|}{0.268}   & 0.415     & 0.293     & \multicolumn{1}{|c}{0.532}   & \multicolumn{1}{c|}{0.371}   & 0.533     & 0.425     \\
\multicolumn{1}{c}{}                                                                         & \multicolumn{1}{c}{}                    & MPR-OB                & \multicolumn{1}{|c}{\textbf{0.447}}   & \multicolumn{1}{c|}{0.299}   & 0.427     & 0.303     & \multicolumn{1}{|c}{\textbf{0.567}}   & \multicolumn{1}{c|}{0.434}   & 0.570     & 0.448    \\
\multicolumn{1}{c}{}                                                                         & \multicolumn{1}{c}{}                    & MPR                   & \multicolumn{1}{|c}{0.445}   & \multicolumn{1}{c|}{\textbf{0.312}}   & \textbf{0.438}     & \textbf{0.309}     & \multicolumn{1}{|c}{0.563}   & \multicolumn{1}{c|}{\textbf{0.440}}   & \textbf{0.576}     & \textbf{0.450}     \\ \cline{2-11} 
\multicolumn{1}{c}{}                                                                         & \multicolumn{1}{c}{\multirow{5}{*}{NY}} & MPR-RP            & \multicolumn{1}{|c}{0.286}   & \multicolumn{1}{c|}{0.241}   & 0.289     & 0.234     & \multicolumn{1}{|c}{0.417}   & \multicolumn{1}{c|}{0.375}   & 0.442     & 0.374     \\
\multicolumn{1}{c}{}                                                                         & \multicolumn{1}{c}{}                    & MPR-FE              & \multicolumn{1}{|c}{0.475}   & \multicolumn{1}{c|}{0.289}   & 0.434     & 0.305     & \multicolumn{1}{|c}{0.550}   & \multicolumn{1}{c|}{0.425}   & 0.615     & 0.426     \\
\multicolumn{1}{c}{}                                                                         & \multicolumn{1}{c}{}                    & MPR-OB                & \multicolumn{1}{|c}{0.480}   & \multicolumn{1}{c|}{0.291}   & \textbf{0.439}     & \textbf{0.305}     & \multicolumn{1}{|c}{0.557}  & \multicolumn{1}{c|}{0.418}   & 0.622     & 0.434     \\
\multicolumn{1}{c}{}                                                                         & \multicolumn{1}{c}{}                    & MPR                   & \multicolumn{1}{|c}{\textbf{0.482}}   & \multicolumn{1}{c|}{\textbf{0.310}}   & 0.429     & 0.301     & \multicolumn{1}{|c}{\textbf{0.569}}   & \multicolumn{1}{c|}{\textbf{0.426}}   & \textbf{0.639}     & \textbf{0.452}     \\ \hline
Dataset
                                              & \multicolumn{1}{l}{}                     & \multicolumn{1}{l}{} & \multicolumn{8}{c}{2020 U.S. Election} \\
                                              Grainularity   & \multicolumn{1}{l}{}                     & \multicolumn{1}{l}{} & \multicolumn{4}{c}{Sentiment}                                     & \multicolumn{4}{c}{Sentiment Polarity}                                      \\
Time &                                          &                       & \multicolumn{2}{c}{T3}                  & \multicolumn{2}{c}{T4} & \multicolumn{2}{c}{T3}                  & \multicolumn{2}{c}{T4} \\ 
                                                                                          Metrics     &                                          &                       & Accuracy & \multicolumn{1}{c}{Macro F1} & Accuracy   & Macro F1   & Accuracy & \multicolumn{1}{c}{Macro F1} & Accuracy   & Macro F1   \\ \hline
\multicolumn{1}{c}{\multirow{4}{*}{\rotatebox{90}{Gemma 2}}}                                               & \multicolumn{1}{c}{\multirow{4}{*}{}} & MPR-RP            & \multicolumn{1}{|c}{0.415}          & \multicolumn{1}{c|}{0.196}   & 0.354     & 0.104     & \multicolumn{1}{|c}{0.477}   & \multicolumn{1}{c|}{0.323}   & 0.391     &  0.173          \\
\multicolumn{1}{c}{}                                                                         & \multicolumn{1}{c}{}                    & MPR-FE              & \multicolumn{1}{|c}{0.516}   & \multicolumn{1}{c|}{0.213}   & 0.407     & 0.190     & \multicolumn{1}{|c}{0.551}   & \multicolumn{1}{c|}{0.356}   & 0.434     & 0.320     \\
\multicolumn{1}{c}{}                                                                         & \multicolumn{1}{c}{}                    & MPR-OB                & \multicolumn{1}{|c}{0.594}   & \multicolumn{1}{c|}{\textbf{0.376}}   & 0.588     & 0.378     & \multicolumn{1}{|c}{0.685}   & \multicolumn{1}{c|}{\textbf{0.521}}   & 0.662     & 0.509     \\
\multicolumn{1}{c}{}                                                                         & \multicolumn{1}{c}{}                    & MPR                   & \multicolumn{1}{|c}{\textbf{0.615}}   & \multicolumn{1}{c|}{0.374}   & \textbf{0.596}     & \textbf{0.397}     & \multicolumn{1}{|c}{\textbf{0.692}}   & \multicolumn{1}{c|}{0.513}   & \textbf{0.669}     & \textbf{0.533}     \\ \cline{2-11} 
\multicolumn{1}{c}{\multirow{4}{*}{\rotatebox{90}{Mistral}}}                                                                         & \multicolumn{1}{c}{\multirow{4}{*}{}} & MPR-RP             & \multicolumn{1}{|c}{0.457}   & \multicolumn{1}{c|}{0.221}   & 0.431     & 0.202     & \multicolumn{1}{|c}{0.567}   & \multicolumn{1}{c|}{0.370}   & 0.588     & 0.352           \\
\multicolumn{1}{c}{}                                                                         & \multicolumn{1}{c}{}                    & MPR-FE              & \multicolumn{1}{|c}{0.500}   & \multicolumn{1}{c|}{0.203}   & 0.404     & 0.193     & \multicolumn{1}{|c}{0.578}   & \multicolumn{1}{c|}{0.325}   & 0.455     & 0.296     \\
\multicolumn{1}{c}{}                                                                         & \multicolumn{1}{c}{}                    & MPR-OB                & \multicolumn{1}{|c}{0.571}   & \multicolumn{1}{c|}{0.362}   & 0.578     & \textbf{0.377}     & \multicolumn{1}{|c}{0.657}   & \multicolumn{1}{c|}{0.486}   & 0.651     & \textbf{0.520}     \\
\multicolumn{1}{c}{}                                                                         & \multicolumn{1}{c}{}                    & MPR                   & \multicolumn{1}{|c}{\textbf{0.593}}   & \multicolumn{1}{c|}{\textbf{0.369}}   & \textbf{0.581}     & 0.370     & \multicolumn{1}{|c}{\textbf{0.668}}   & \multicolumn{1}{c|}{\textbf{0.495}}   & \textbf{0.654}     & 0.508     \\ \hline
\end{tabular}
}
\end{table*}

\subsection{Ablation Studies}
To demonstrate the effectiveness of our framework, we conducted extensive ablation studies using different LLMs. As shown in Table \ref{tab: ablation}, we performed three sets of experiments, each removing a key component from the full framework. The "MRP-RP" configuration directly forecasts the sentiment score during role-playing, removing the module to predict the next social media comment. "MRP-FE" removes the feature extraction module and directly predicts the social media comment through role-play. "MRP-OB" removes the objective fine-tuned "psychology" LLM, where role-playing is conducted solely based on the extracted features of the users.

Our proposed framework consistently outperforms the variants without important modules. The MRP-RP variant, comparable to random guessing, reflects the challenges humans face when predicting sentiment scores without proper context simulation. This result highlights the importance of replicating human action and thought processes in LLMs for effective human-oriented studies. Directly predicting sentences without feature extraction (MRP-FE) introduces high stochasticity. The \redtext{MRP-OB} variant, while slightly less effective than the full framework, underscores the value of the fine-tuned behavioral psychologist LLM, which was refined with just 25,000 Q\&A instances. For applications focusing solely on sentiment polarity, our proposed framework can achieve an accuracy rate as high as 63.9\%, demonstrating its capability to comprehend, reason, and forecast sentiment for social media users. 

\redtext{
\subsection{Discussions}
We study the difference between the performance of both datasets and the error cases. Even though the performance improved substantially, there are still potential improvements. First, in our experiments, the subjective role-playing agents only have access to limited information compared to real-life users who have diverse information sources like friends, families, local news, etc. Future comprehensive studies might require access to diverse sources to better simulate the user's information gain. Moreover, some comments are stochastic, which is hard to predict. In the case of Hurricane Sandy, more than 15\% of people posted about their circumstances enduring Sandy without relying on information on social media, which makes their emotions hard to predict. On the other hand, the topic of the U.S. election dataset relies more extensively on news and social media output, which partially contributes to the better performance.}

The study of sentiment forecasting has a variety of potential applications. For large-scale natural disaster events, it could be used to detect and predict the areas with the worst mental conditions. In large social events, detecting and predicting extreme sentiment and emotion could help prevent potential chaos. In finance, our framework could be adapted to analyze finance-related social media comments to infer underlying market trends and correlations in a timely manner. Moreover, the ability to interpret and infer future sentiment is also crucial in the development of artificial intelligence with the ability to understand of theory of mind. An artificial agent with such an ability would also help various labor-intensive customer service scenarios. 
\section{Conclusion}
In this paper, we target the problem of sentiment forecasting in social media to predict a user's future sentiment towards a given event. We proposed the context-aware multi-perspective role-playing framework to integrate the social media information up to time $t$ to predict the sentiment at time $t' \ge t$. Experiments show that our proposed framework outperforms the state-of-the-art methods on both macroscopic and microscopic levels. The implementation is currently available at \url{https://github.com/ManFanhang/Context-Aware-Sentiment-Forecasting-via-LLM-based-Multi-Perspective-Role-Playing-Agents}.

\section*{Limitations}
First, many widely used base models, such as the GPT series, are explicitly trained to avoid processing or generating social media content with inappropriate or strongly negative tones. Consequently, our proposed framework is implemented with LLMs that are not constrained by such politeness or political correctness filters. The performance varies depending on the choice of LLM, highlighting a dependency on model architecture and training objectives. The proposed method may restrict its applicability to certain off-the-shelf models due to the restrictions. A sample performance of the sanctioned model Llama 3.1 can be found in Table \ref{tab: ablation-llama} in the Appendix. Furthermore, the current framework merely includes textual data, which captures only a subset of the information available on social media platforms. With more and more information expressed through different modalities, future work could explore multimodal integration to enable more comprehensive sentiment forecasting.

\bibliography{acl_latex}
\appendix
\section{Appendix}

\subsection{Instruction Demonstration}\label{app: instruction demonstration}
\begin{figure*}[htbp]
  \centering
  \includegraphics[width=\textwidth]{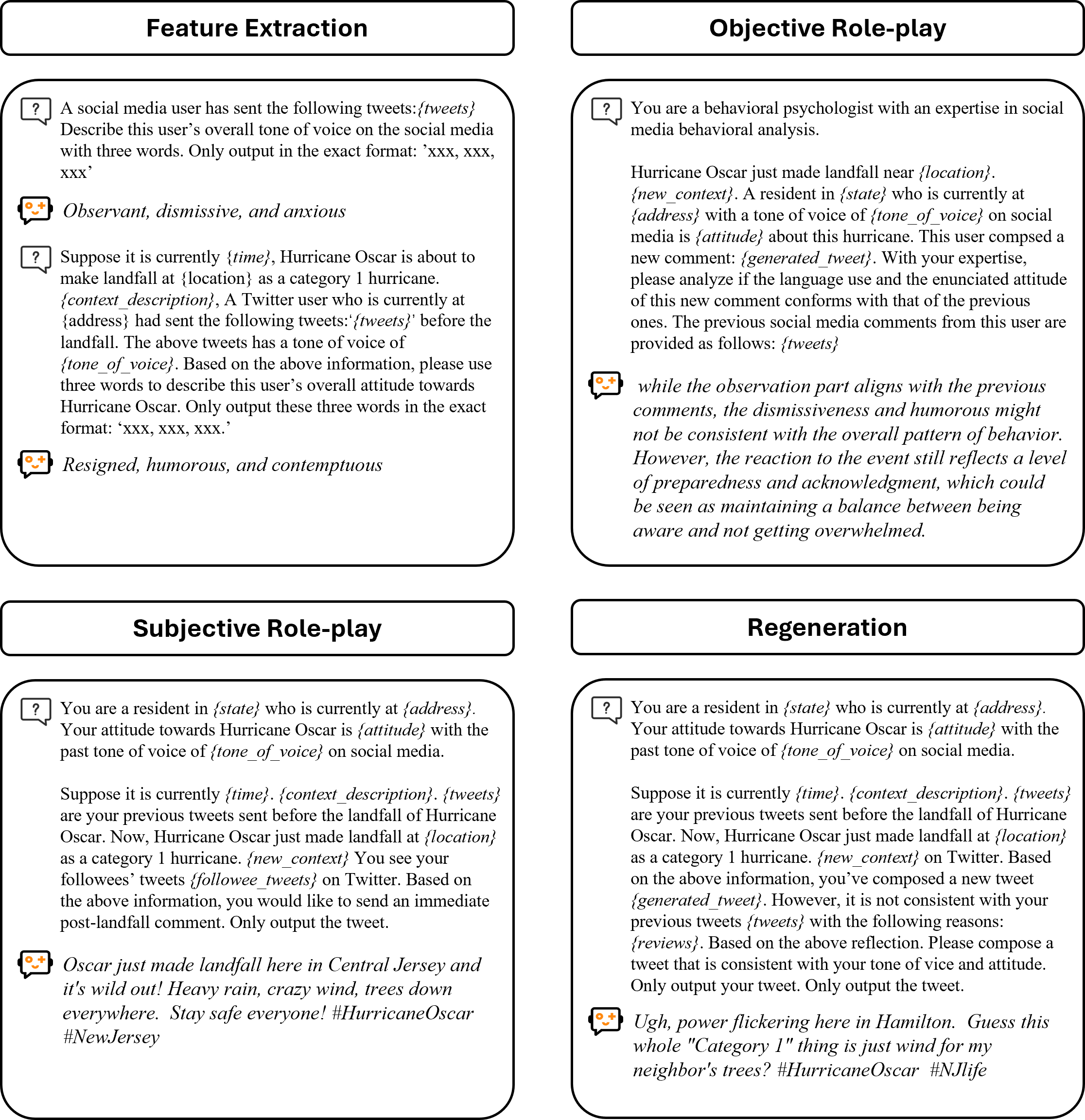} 
  \caption{The demonstration of our constructed instructions for feature extraction, subjective role-play, and objective role-play.}
  \label{fig: demo}
\end{figure*}

Our proposed instruction and sample results are demonstrated in Figure \ref{fig: demo}. We took an social media user in New Jersey as an example. The goal is to predict the sentiment of this user immediately after the landfall. Since the LLMs are trained with a massive collection of corpora data, they contain retrospective knowledge of Hurricane Sandy. To verify the ability to predict the sentiment of a large crowd, the instruction should be designed to omit the effect of retrospective knowledge by constructing a hypothetically identical environment as if it were in a parallel universe. With the instruction, we generated only one social media content. Pre-experiment shows that multiple generations with Gemma 2 9B and Mistral Nemo 12B under the same prompt result in a 4.2\% differ in the sentiment of the generated comment which is not statistically signnificant. 

\subsection{Detailed Procedure for Fine-Tuning Llama3 8B with LoRA}\label{app: fine tuning}
The fine-tuning process was carried out using the Llama3 8B Instruct model. The embedding layer includes \texttt{embed\_tokens} with dimensions (128256, 4096). The decoder comprises 32 \texttt{LlamaDecoderLayer} instances, each with self-attention (\texttt{LlamaSdpaAttention}), including \texttt{q\_proj}, \texttt{k\_proj}, \texttt{v\_proj}, \texttt{o\_proj}, and \texttt{rotary\_emb}, as well as an MLP (\texttt{LlamaMLP}) with \texttt{gate\_proj}, \texttt{up\_proj}, \texttt{down\_proj}, and activation function SiLU. Additionally, it includes \texttt{input\_layernorm} and \texttt{post\_attention\_layernorm} layers, and a final normalization layer (\texttt{LlamaRMSNorm}). The output layer is a linear transformation (\texttt{lm\_head}) from 4096 to 128256 dimensions. The model uses \texttt{torch.bfloat16} data type for processing.

The LoRA configuration was crucial for our fine-tuning approach. The configuration parameters include \texttt{LoraConfig} with \texttt{peft\_type} set to LORA, \texttt{task\_type} as CAUSAL\_LM, and \texttt{r} value of 8. The target modules affected by LoRA are \texttt{down\_proj}, \texttt{v\_proj}, \texttt{up\_proj}, \texttt{q\_proj}, \texttt{k\_proj}, \texttt{o\_proj}, and \texttt{gate\_proj}, with \texttt{lora\_alpha} set to 32 and \texttt{lora\_dropout} at 0.1. The model was configured with 20,971,520 trainable parameters out of 8,051,232,768 parameters, making the trainable percentage approximately 0.26\%. 

\begin{figure*}
    \centering
    \includegraphics[width=\linewidth]{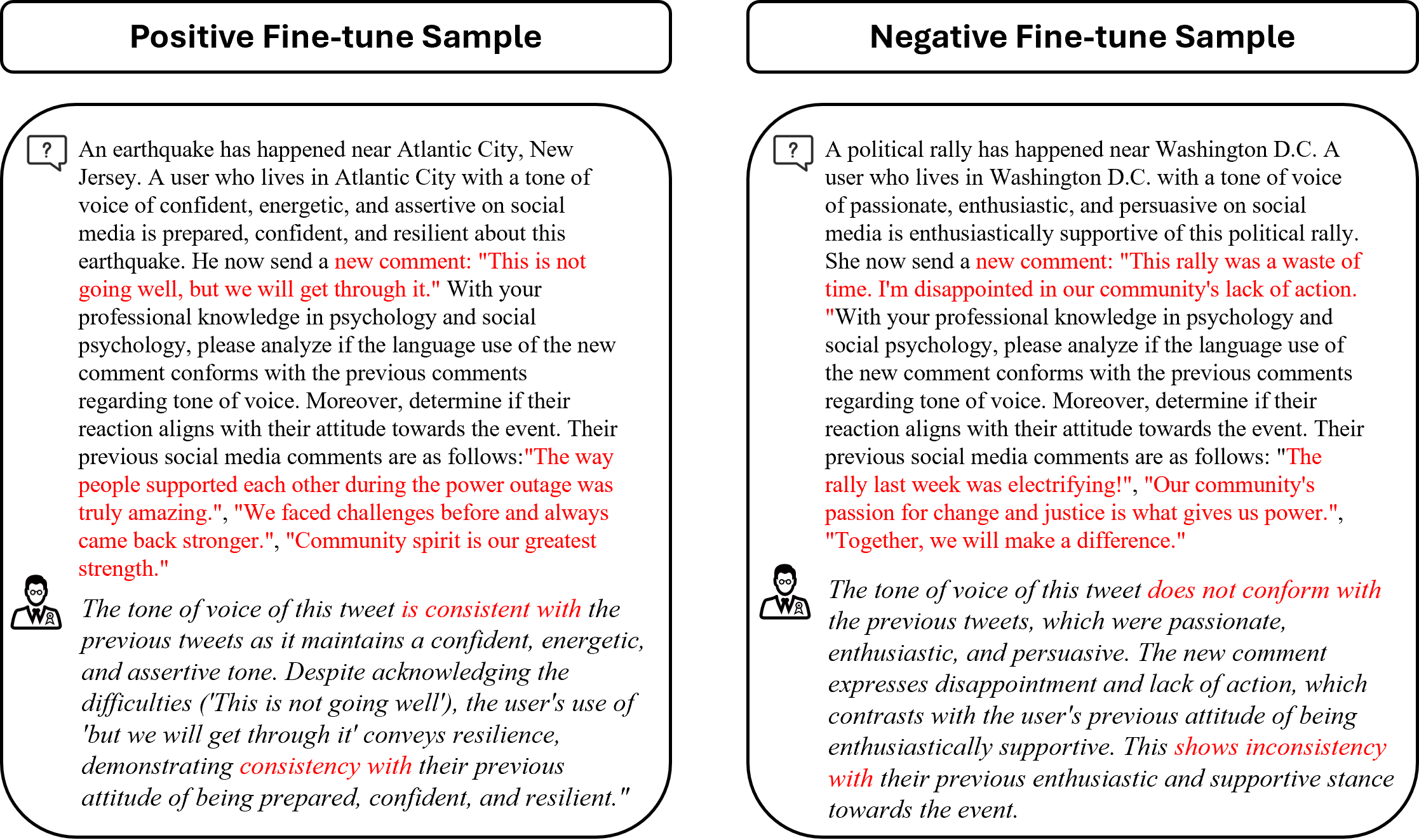}
    \caption{The expert example of the tone of voice and attitude consistent analysis.}
    \label{fig: expert}
\end{figure*}

\begin{figure}
    \centering
    \includegraphics[width=\linewidth]{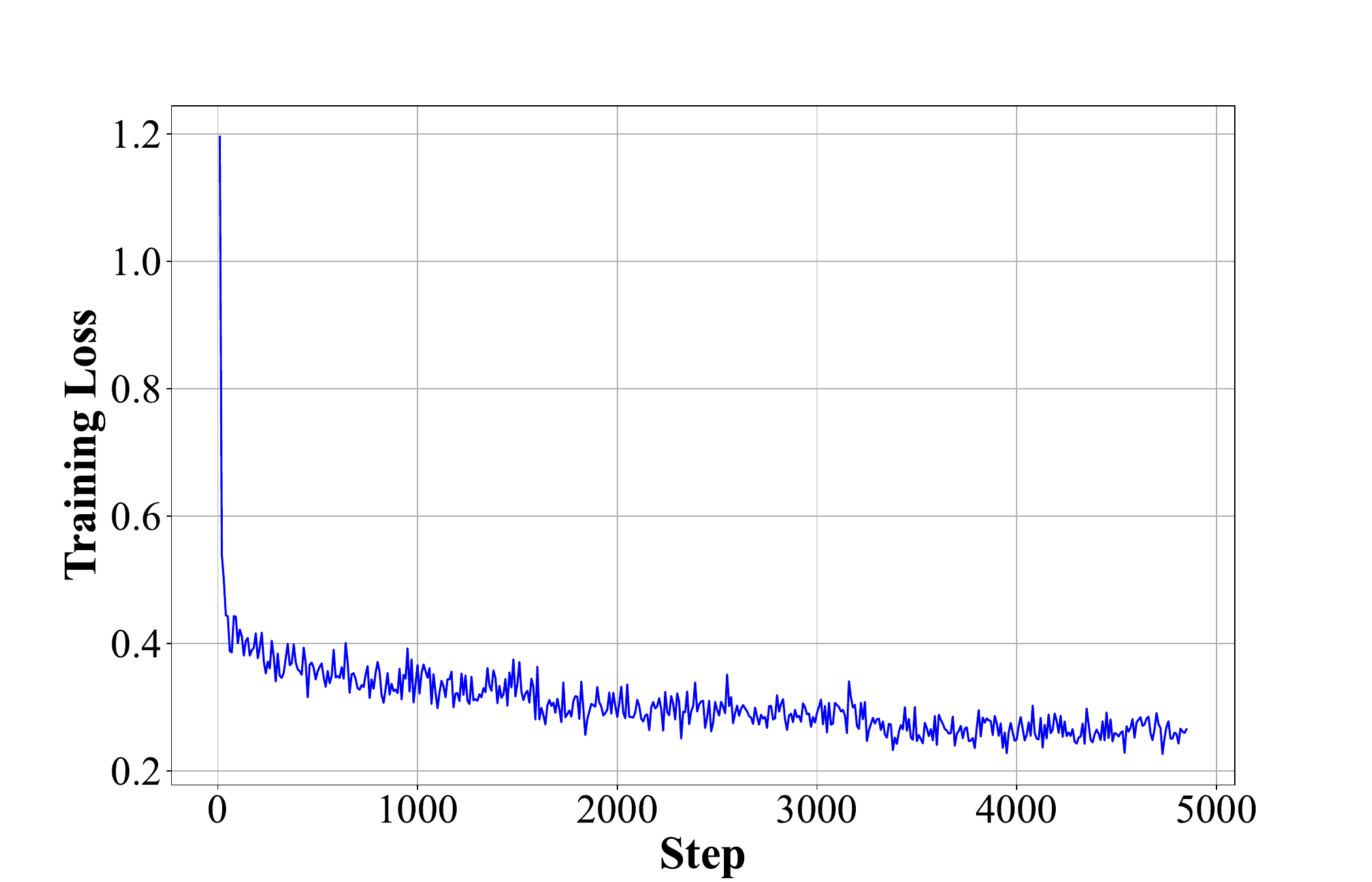}
    \caption{Fine-tune training loss for Llama 3 8b Instruct.}
    \label{fig: loss}
\end{figure}

The training loss over the steps is visualized in \ref{fig: loss}, which illustrates the progressive decrease in loss, indicating the model's improvement over the training period. This appendix provides a detailed description of the Llama3 8B Instruct model architecture, configuration, and training process utilizing the LoRA fine-tuning method. The visualization of the training loss demonstrates the model's convergence and the effectiveness of the fine-tuning approach. The sample Q\&A pairs for fine-tuning are shown in Figure~\ref{fig: expert}.

\subsection{Event Context with Timeline} \label{app: summary}

\textbf{Hurricane Sandy} is one of the most destructive and widely recognized disasters in U.S. history \cite{kryvasheyeu2015performance}. The corresponding dataset comprises a comprehensive collection of Twitter messages from Oct. 15 to Nov. 12, 2012, spanning the period a week before the hurricane's formation and ten days after its dissipation. The metadata includes followee and friend counts, retweet statuses, follower-followee relationships, locations (self-reported or automatically detected), and timestamps. It comprises a total of 52.55 million messages from 13.75 million unique users, offering a detailed view of public sentiment and communication patterns during this significant natural disaster. \redtext{Hurricane Sandy was formed south-west of Kingston, Jamaica on Oct. 22, 2012. It made landfall in Jamaica as a C1 hurricane on Oct. 24 and in Cuba as a C3 hurricane 10 hours later. On Oct. 29, Hurricane Sandy hit Brigantine, New Jersey as a C1 hurricane. The storm surge was as high as 3.85m, with prevalent levels between 0.8 and 2.6m along the coast of New Jersey and New York. New Jersey bore the brunt of the storm. New York, particularly Long Island, was heavily affected due to its terrain, but was rather farther from the storm's center. We mainly chose densely populated areas from New York and strongly affected areas from New Jersey to conduct the experiments. By Nov. 5, Hurricane Sandy was no longer active but had transitioned into a post-tropical cyclone. The government strived to perform post-disaster relief. However, many areas are still without power.}  

\redtext{\textbf{U.S. presidential election} was one of the most influential events. The second dataset corresponds to the 2020 U.S. presidential election, where both candidates directly used Twitter to help express themselves and political opinions ~\cite{kaggle2020electiontweets,caballero2021predicting}. It contains 1.7 million tweets spanning from Oct. 15 to Nov. 8, 2020, the period of the climax of the presidential race. The metadata contains users' descriptions of themselves, followee and friend counts, locations, and timestamps. On October 22, 2020, the final presidential debate between Donald Trump and Joe Biden took place in Nashville, Tennessee, focusing on various policy issues. On Nov. 3, 2020, Election Day saw a historic voter turnout, with both in-person and absentee voting contributing to the largest voter participation in over a century. The votes were counted from Nov. 4 to Nov. 7. However, the key states (swing states) experienced delays due to absentee votes. On Nov.7, Major news outlets projected Joe Biden as the winner of the election after Pennsylvania's results gave him enough electoral votes. However, President Trump contested the results. Claim of vote fraud.}


\subsection{Preprocessing} \label{app: preprocessing}
Our reasoning problem focuses on generating the next OSN comment for a user. However, corpora related to Hurricane Sandy were used to train the LLMs. Referring to Hurricane Sandy could lead to the retrieval of posterior knowledge, compromising the integrity of a prospective study. To avoid this interference, we renamed the event from Hurricane Sandy to Hurricane Oscar, a candidate hurricane name for 2024, and adjusted the timeline to 2024. This ensures that knowledge about Hurricane Sandy is vaguely correlated. For the Hurricane Sandy dataset, keywords and hashtags such as "Hurricane," "Sandy," "storm," "power," etc. For the US 2020 election dataset, keywords and hashtags included "presidential election," "Biden," "Trump," etc. Noted that the context of the event, $\mathcal{E}$, captures the broader context of the event, which includes background knowledge, news updates, and evolving information surrounding the event. The source spans from official news websites, TV, and official announcements. On the other hand, the Followees comments, $\mathcal{F}$, consist of the posts and social media outputs from the followees, which are filtered and analyzed within our experiments. If a user $u$ follows an official news account, it would also be formalized as $\mathcal{F}^u$. $\mathcal{E}^u$ and $\mathcal{F}^u$ could have overlapping information.

Moreover, we mapped OSN comments to a unified sentiment metric. Sentiment analysis is a complex field, encompassing lexicon-based, machine learning-based, and, most recently, LLM-based methods. Although LLMs can perform sentiment analysis on OSN comments, they require extensive resources and can exhibit high stochasticity~\cite{lakhanpal2023leveraging}. The same instruction might produce different sentiment labels for the same sentence, introducing instability into the proposed methods. 
After careful consideration, we selected the state-of-the-art supervised learning model bert-base-multilingual-uncased-sentiment.

\subsubsection{Ground Truth}
We first used this model to perform text-sentiment mapping on corpora of the original dataset, establishing the ground truth. Our problem is designed to predict the future expressed sentiments on OSN. We can either choose the next exact tweet or a collection of OSN users given a certain timeframe to make up the ground truth. The first scheme involves directly selecting the subsequent message from each user as the ground truth for scoring. Conversely, the second scheme computes the average value of all messages from each user during the specified period, in our case, 24 hours. 

To verify the consistency between two distinct design schemes for determining ground truth in our study, we employed the Kolmogorov-Smirnov (K-S) test. The K-S test results indicated a K-S Statistic of 0.0045, suggesting a minimal distributional difference between the two samples. Additionally, the P-value was found to be 0.9999, which implies that there is no significant difference between the distributions of the two schemes.

Based on these findings, we opted to utilize the second method for characterizing the ground truth. The rationale behind this choice lies in the stability of the second scheme. By averaging all messages from each user over the given period, this method mitigates potential anomalies and provides a more robust representation of user behavior.

\subsubsection{User Selection}
Then, we applied the model to analyze the sentiment of our OSN comments, ensuring uniformity across experiments. The model assigns a discrete sentiment score ranging from -2 to 2, where -2 indicates strongly negative sentiment and 2 indicates strongly positive sentiment. This five-category sentiment score allows for a fine-grained analysis of OSN user sentiments, capturing subtle variations in public sentiment.

To focus on users who freely express themselves on OSNs, we carefully selected user samples, omitting official accounts and news outlets. The experiment is conducted with 3000 users from New Jersey and 3000 users from New York, distributed in 15 Counties.
Social media behavior is complex and subject to complex influence factors, especially for internet celebrities, the government, and the news. The study of crowd sentiment prediction, on the other hand, focuses on the intuitive and reactive response of the general netizens. The user should not be socially influential but rather influenced by OSN. We should exclude bots, news outlets, internet celebrities, public figures, and government voices. Therefore, we strategically select the users to represent the general netizens based on the following criteria.
\begin{enumerate}
  \item The user's followees are accessible, indicating active interaction and engagement within the community.
  \item The user's tweet history should contain mentions of "sandy" or "hurricane" to prevent irrelevance.
  \item The threshold for total tweet counts is set to range from 10 to 1000, filtering out extremely inactive and active users.
  \item The user's followers count and friends count were restricted to between 100 and 2000, to avoid unusually high or low social influences.
  \item The presence of a geographic label was required to perform spatial segmentation for the user crowd. To perform geographical studies, we carefully select users from 15 different counties across New York and New Jersey.
  \item The user must have comments before, during, and a week after the emergency to form a comprehensive view and the evolution of user sentiment.
\end{enumerate}

There are a total of 124,876 users satisfying these criteria with a total of 5,038,920 tweets. For large enough users, we could assume that the distribution of their behaviors. We group the users based on their geographical label. A region of the directly affected area was defined based on the trajectory and the diameter of the wind cycle of Hurricane Sandy. The users were segmented to be either directly affected or not affected. Moreover, to predict the evolution of user sentiment, we made predictions during and after the landfall.

\begin{table*}[!t]
\caption{The extended ablation study performed with aligned model Llama 3,1. Bold denotes the best (highest) results}
\label{tab: ablation-llama}
\resizebox{\textwidth}{!}{%
\begin{tabular}{ccccccccccc}
\hline
                                                                                               & \multicolumn{1}{l}{}                     & \multicolumn{1}{l}{} & \multicolumn{4}{c}{Sentiment}                                     & \multicolumn{4}{c}{Sentiment Polarity}                                      \\ 
                                                                                               &                                          &                       & \multicolumn{2}{c}{T1}                  & \multicolumn{2}{c}{T2} & \multicolumn{2}{c}{T1}                  & \multicolumn{2}{c}{T2} \\ 
                                                                                               &                                          &                       & Accuracy & \multicolumn{1}{c}{Macro F1} & Accuracy   & Macro F1   & Accuracy & \multicolumn{1}{c}{Macro F1} & Accuracy   & Macro F1   \\ \hline
\multicolumn{1}{c}{\multirow{8}{*}{\rotatebox{90}{Llama 3.1}}}                                               & \multicolumn{1}{c}{\multirow{4}{*}{NJ}} & MPR-RP            & \multicolumn{1}{|c}{0.150}          & \multicolumn{1}{c|}{0.150}   & 0.127     & 0.106     & \multicolumn{1}{|c}{0.185}   & \multicolumn{1}{c|}{0.150}   & 0.254     &  0.174          \\
\multicolumn{1}{c}{}                                                                         & \multicolumn{1}{c}{}                    & MPR-FE              & \multicolumn{1}{|c}{0.192}   & \multicolumn{1}{c|}{0.115}   & 0.308     & 0.159     & \multicolumn{1}{|c}{0.243}   & \multicolumn{1}{c|}{0.165}   & 0.422     & 0.247     \\
\multicolumn{1}{c}{}                                                                         & \multicolumn{1}{c}{}                    & MPR-OB                & \multicolumn{1}{|c}{\textbf{0.184}}   & \multicolumn{1}{c|}{\textbf{0.113}}   & \textbf{0.312}     & 0.161     & \multicolumn{1}{|c}{0.240}   & \multicolumn{1}{c|}{0.168}   & 0.424     & 0.151     \\
\multicolumn{1}{c}{}                                                                         & \multicolumn{1}{c}{}                    & MPR                   & \multicolumn{1}{|c}{0.173}   & \multicolumn{1}{c|}{0.108}   & 0.308     & \textbf{0.163}     & \multicolumn{1}{|c}{\textbf{0.255}}   & \multicolumn{1}{c|}{\textbf{0.173}}   & \textbf{0.425}     & \textbf{0.247}     \\ \cline{2-11} 
\multicolumn{1}{c}{}                                                                         & \multicolumn{1}{c}{\multirow{4}{*}{NY}} & MPR-RP             & \multicolumn{1}{|c}{0.137}   & \multicolumn{1}{c|}{0.143}   & 0.197     & 0.148     & \multicolumn{1}{|c}{0.138}   & \multicolumn{1}{c|}{0.113}   & 0.267     & 0.204           \\
\multicolumn{1}{c}{}                                                                         & \multicolumn{1}{c}{}                    & MPR-FE              & \multicolumn{1}{|c}{\textbf{0.178}}   & \multicolumn{1}{c|}{0.098}   & 0.250     & 0.141     & \multicolumn{1}{|c}{0.223}   & \multicolumn{1}{c|}{0.148}   & 0.337     & 0.216     \\
\multicolumn{1}{c}{}                                                                         & \multicolumn{1}{c}{}                    & MPR-OB                & \multicolumn{1}{|c}{0.172}   & \multicolumn{1}{c|}{0.101}   & \textbf{0.254}     & \textbf{0.154}     & \multicolumn{1}{|c}{0.226}   & \multicolumn{1}{c|}{0.160}   & \textbf{0.342}     & 0.217     \\
\multicolumn{1}{c}{}                                                                         & \multicolumn{1}{c}{}                    & MPR                   & \multicolumn{1}{|c}{0.158}   & \multicolumn{1}{c|}{\textbf{0.103}}   & 0.243     & 0.152     & \multicolumn{1}{|c}{\textbf{0.227}}   & \multicolumn{1}{c|}{\textbf{0.163}}   & 0.340     & \textbf{0.218}     \\ \hline
\end{tabular}
}
\end{table*}

\subsection{Performance Analysis for Different LLMs}
Due to freedom of speech, social media content may include swear words and politically incorrect expressions. Some popular LLMs, including the GPT series from OpenAI and the Llama series from Meta, are tuned to omit processing and generation of such content~\cite{achiam2023gpt}. Instead of conducting the analysis, these LLMs would say that they can't proceed with such content. When we tried to use GPT and Llama to perform an analysis of social media comments, the tone of voice and attitude for aggressive and politically incorrect phrases could not be analyzed and the social media comment generation failed due to censorship aimed at aligning with human values~\cite{pletenev2024somethingawful}. When we try to perform feature extraction and subjective role-play to generate the post-landfall social media comments, 1547 out of 3000 social media users from New Jersey and 1633 out of 3000 social media users from New York are detected to have aggressive expressions. For the same 6,000 users, llama avoided generating social media comments for 468 users from New Jersey and 854 New Jersey users. We treat these results as false for all categories, significantly decreasing the accuracy rate and the F1 score as shown in Table \ref{tab: ablation-llama}, the extended ablation study with LLama 3,1.  

We used the bert-base-multilingual-uncased-sentiment model to perform sentiment analysis for the tweets detected to be politically incorrect or toxic. Over 82\% is assigned with the score $-2$ (strongly negative), and 94\% are detected to be negative. It aligns with our assumption that the aggressive comments are mostly strongly negative. On the other hand, most politically incorrect tweets with positive sentiments use some swear words or politically incorrect terms as slang or exclamation. 

\end{document}